\documentclass[conference]{IEEEtran}

\IEEEoverridecommandlockouts
\usepackage{cite}
\usepackage{amsmath,amssymb,amsfonts}
\usepackage{algorithm,algpseudocode}
\usepackage{cellspace}
\usepackage{graphicx}
\usepackage{textcomp}
\usepackage{soul}
\usepackage{multirow}
\usepackage{comment}
\usepackage{xcolor}
\usepackage{tcolorbox}

\usepackage{float}

\usepackage[compatibility=false]{caption}
\usepackage{subcaption}
\usepackage{hyperref}
\usepackage{booktabs}
\usepackage{cellspace}
\usepackage{siunitx}

\def\BibTeX{{\rm B\kern-.05em{\sc i\kern-.025em b}\kern-.08em
    T\kern-.1667em\lower.7ex\hbox{E}\kern-.125emX}}
\begin{document}

\title{OmniPath: A Multi-Modal Agentic Framework for Auditing Wheelchair Accessibility}




\author{
\IEEEauthorblockN{
ASM Mobarak Hossain\IEEEauthorrefmark{1},
Nadim Mahmud\IEEEauthorrefmark{2},
Vaskar Raychoudhury\IEEEauthorrefmark{2},
Md Osman Gani\IEEEauthorrefmark{1}
}
\IEEEauthorblockA{
\IEEEauthorrefmark{1} Causal AI Lab, Department of Information Systems,
University of Maryland Baltimore County, Baltimore, MD, USA\\
\IEEEauthorrefmark{2} Department of Computer Science \& Software Engineering,
Miami University, Oxford, OH, USA\\
Email: \{ir20494,mogani\}@umbc.edu,
\{mahmudm2,raychov\}@miamioh.edu
}
}

\maketitle

\begin{abstract}
For a wheelchair user, a standard blue line on a map is often a broken promise. While platforms like OpenStreetMap (OSM) successfully capture where a path is, they frequently fail to convey how it physically feels to travel on it. This information barrier is problematic for wheelchair users. To solve this issue, we present OmniPath, a system that moves from passive mapping to proactive environmental auditing. Our framework fuses the network topology of OSM with the submeter precision of high-density aerial LiDAR (USGS 3DEP) to create a high-fidelity 3D model of the pedestrian environment. Rather than simply routing a user, our agent virtually traverses the network, analyzing the surface in 0.5 meter increments. It rigorously quantifies physical friction points specifically running slope, cross slope, and vertical discontinuities against ADA compliance standards, calculating a weighted severity score to categorize hazards from ``Mild'' to ``Critical.'' To ensure real world reliability, we validated the system against 200 physical ground truth field surveys across the National Mall using stratified random sampling. The framework demonstrated strong diagnostic reliability for high-severity hazards, achieving F1-scores of 0.60 for Severe and 0.58 for critical categories. By automating this micro-scale inspection, OmniPath identifies the ``invisible'' barriers that standard maps miss, effectively transforming a static dataset into accessibility data source that anticipates accessibility challenges before the user ever leaves home.
\end{abstract}


\section{Introduction}
\label{sec:intro}


\begin{figure*}[htp] 
    \begin{subfigure}[b]{0.38\textwidth}
      \centering
      \includegraphics[width=\textwidth, height=4cm]{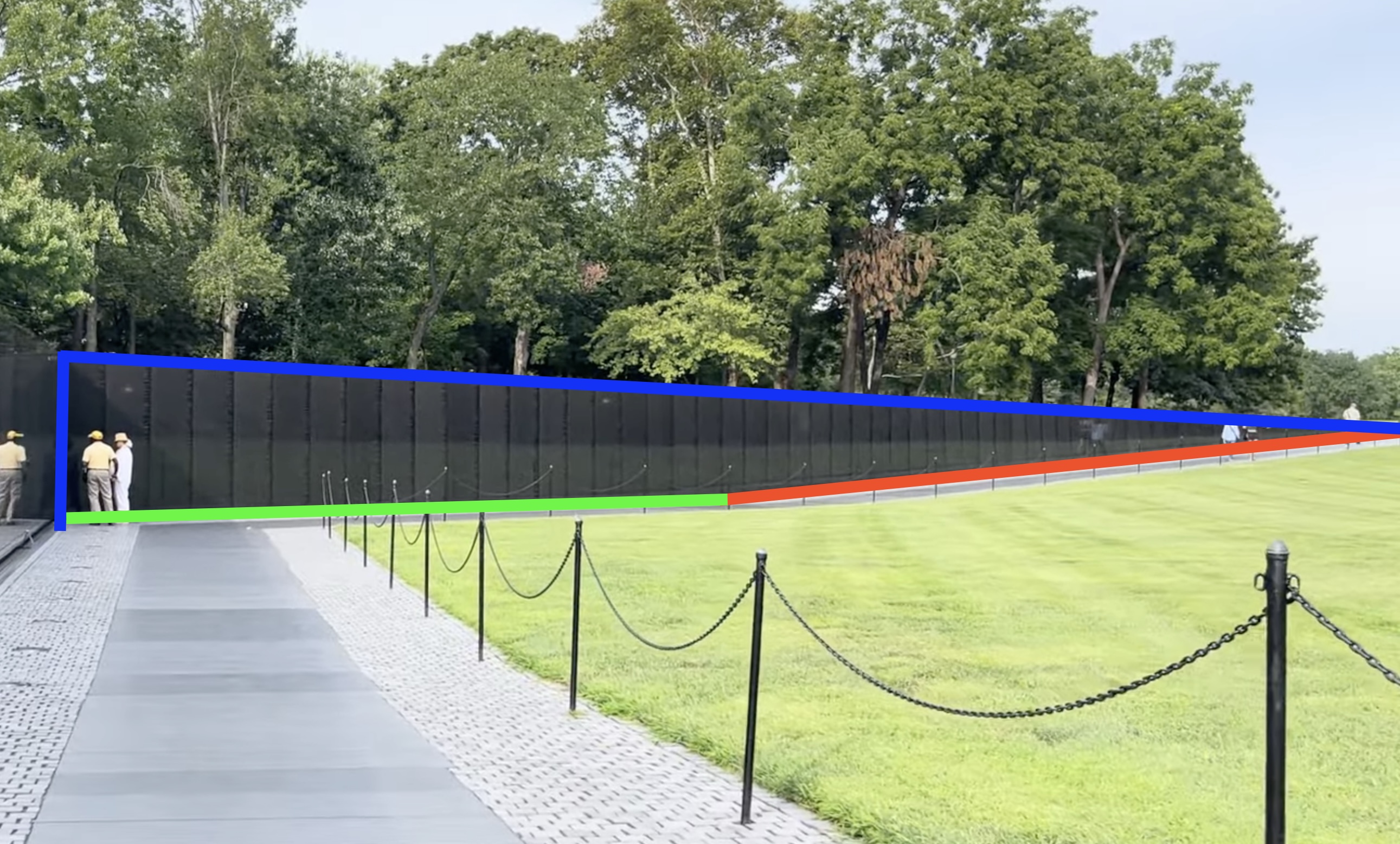}
      \caption{Inaccessible Slope (Red line)}
    \end{subfigure}\hfill
    \begin{subfigure}[b]{0.26\textwidth}
      \centering
      \includegraphics[width=\textwidth, height=4cm]{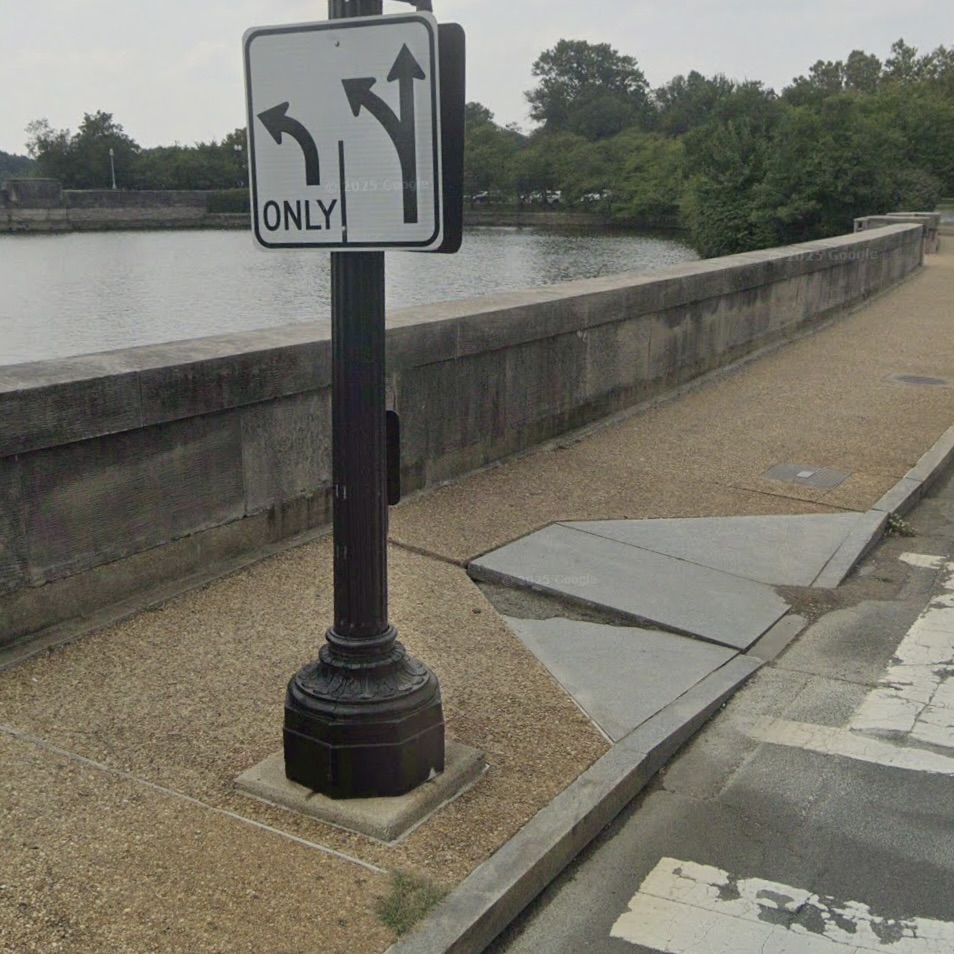}
      \caption{Obstacle on Sidewalk}
    \end{subfigure}\hfill
    \begin{subfigure}[b]{0.26\textwidth}
      \centering
      \includegraphics[width=\textwidth, height=4cm]{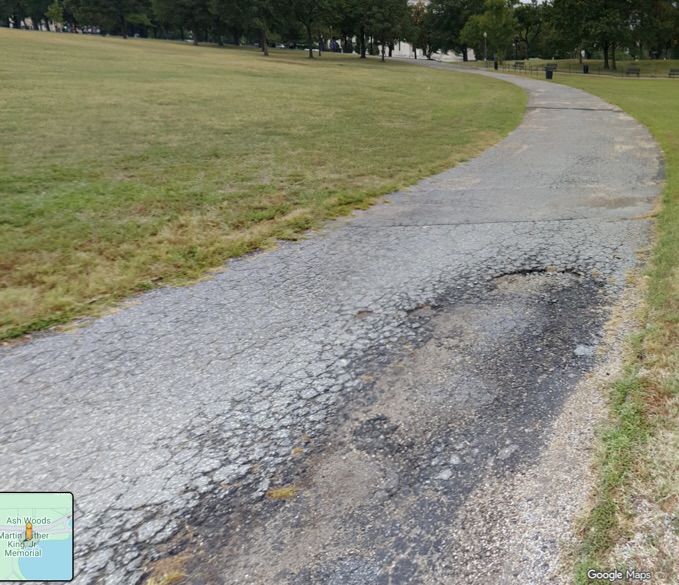}
      \caption{Potholes}
    \end{subfigure}
    \caption{Accessibility barriers detected by OmniPath: (a) excessive running slope (11.76\%, 6.7$^{\circ}$) at the Vietnam Veterans Memorial; (b) excessive running slope (20.81\%, 11.76$^{\circ}$) and cross slope (12.99\%, 7.4$^{\circ}$) along Independence Avenue SW; (c) 24 ft (733 cm) pavement discontinuity at West Potomac Park, Washington, D.C., USA. (Sources: online)}
    \label{fig:non-accessible_sidewalks}
\end{figure*} 

The challenge of generating optimal navigation routes has been extensively solved for vehicular traffic and able-bodied pedestrians. Modern routing engines prioritize the shortest distance or fastest time, assuming a relatively homogeneous travel surface. However, this ``efficiency-first'' paradigm frequently fails the 65 million wheelchair users globally~\cite{WHO2008_wheelchair_guidelines}, for whom the physical quality of a path is vastly more critical than its length. As documented in our field studies (see Figure 1), what appears to be a valid path on a standard digital map is often rendered entirely inaccessible in the physical world due to ``micro-barriers'' such as broken pavement, steep inclines ($>6^\circ$), or deep vertical surface gaps.

While regulations like the Americans with Disabilities Act (ADA) \cite{ADA} mandate that public paths remain stable, firm, and slip-resistant, the reality of urban infrastructure drastically diverges from these legal prescriptions. Conventional routing systems remain largely reactive to these discrepancies. They rely heavily on retrospective user reports (e.g., Wheelmap \cite{WheelMap}, crowdsourcing) or generic sensor data that fails to capture granular surface fidelity. Existing works \cite{zambonelli2011pervasive, magus} have made significant strides in using crowdsourced data to warn users of obstacles. However, these human-in-the-loop models suffer from inherent data sparsity; if a previous user has not physically traversed and reported a specific barrier, the system remains blind to its dangers, forcing the next user to discover it the hard way.

A critical limitation of current navigation systems is their foundational reliance on municipal vector datasets, which prioritize network connectivity over physical accessibility. A stark example of this failure is found within the official DC GIS ``Sidewalks'' dataset \cite{dc_sidewalks_2023}. To maintain topological continuity for general pedestrians, this dataset explicitly encodes ``any stair feature less than 5 stairs'' as a standard sidewalk. For an able-bodied pedestrian, a three-step staircase is a negligible variation in a walking path. For a manual wheelchair user, however, it is an impassable wall. 

This semantic generalization creates dangerous \textit{Accessibility Traps} routes that appear perfectly valid on a digital map but are physically impossible to traverse. A routing engine relying solely on unimodal vector data would confidently direct a wheelchair user down a path that terminates in a flight of four stairs, forcing an exhausting and hazardous backtrack. This highlights a fundamental truth of accessible navigation: while a 2D GIS layer provides the \textit{intent} of a path, only a 3D topographical layer can validate the \textit{reality} of its surface.

To bridge this gap and dismantle these accessibility traps, we must migrate from reactive mapping to proactive agentic systems. We introduce \textbf{OmniPath}, an Agentic AI framework that transforms the navigation problem from simple pathfinding into complex environmental negotiation.  Instead of waiting for a user to encounter a barrier, OmniPath proactively audits the environment to predict accessibility friction before a route is ever requested. 

OmniPath extends our prior MyPath system \cite{gani2019smart, yu2022automated, nguyen2023mypath, hossain2025smart} and fuses OpenStreetMap (OSM) \cite{OpenStreetMap} network topology with high-resolution aerial LiDAR point clouds, we deploy a virtual auditing agent that traverses the graph, analyzing surface fidelity in 0.5-meter sliding windows. This allows the system to detect distinct physical violations such as cross slopes exceeding 2\% or vertical discontinuities greater than 20 mm that are entirely invisible to standard satellite imagery and 2D maps.

The specific contributions of this paper are as follows:

\begin{itemize}
    \item \textbf{Proactive Agentic Auditing Framework:} We propose a paradigm shift from passive data collection to autonomous environmental auditing. Rather than relying on human user reports, our system proactively scans high-fidelity LiDAR datasets to assign an Accessibility Impedance Score to every network segment, predicting micro-barriers without requiring a physical traverse.
    \item \textbf{Multi-Modal Fusion of OSM and LiDAR:} We introduce a robust technical methodology that fuses 2D vector topology with 3D aerial topography. This fusion enables the extraction of micro-scale road features at a granularity (0.5-meter intervals) previously unattainable by GPS-based smartphone sensing alone.
    \item \textbf{ADA-Compliant Anomaly Detection:} We present an algorithmic approach to rigorously validate paths against specific ADA prescriptions. By automatically identifying and classifying non-compliant geometric features (e.g., drift-inducing cross slopes or entrapment hazards), OmniPath provides a verifiable, objective safety metric for accessible route generation.
\end{itemize}

\section{Related Work}

Navigating urban environments remains challenging for wheelchair users due to inaccessible curbs, uneven surfaces, missing sidewalk segments, and insufficiently mapped pedestrian infrastructure. Although platforms such as OpenStreetMap (OSM) provide valuable geographic data, accessibility-related attributes are often missing. High-resolution geospatial datasets, such as LiDAR collected by the U.S. Geological Survey (USGS)~\cite{usgs_lidar_explorer_2026, usgs_downloader_2026}, remain underutilized for identifying and updating accessibility-relevant infrastructure.
 
Ugalde et al.\ developed a GIS-based ramp accessibility reporting tool to improve urban mobility for people with disabilities~\cite{ugalde2022barrier, silva2023accessibility, grigioni2024safe}. Das et al.\ proposed eNav, a multi-layer routing framework integrating OSM data, airborne laser scanning, sensor-based surface assessment, crowdsourced feedback, and real-time transit information~\cite{dasenhancement, polenakis2024towards, sahoo2024autonomous}. Misra et al.\ developed AdaGen, a deep learning framework for transferring surface classification knowledge between manual and power wheelchairs~\cite{misra2024adagen, ramaraj2024development}, enabling shortest and energy-efficient route selection that accommodates electric-powered wheelchair (EPW) constraints such as battery usage and slope.
 
Several studies have applied computer vision to satellite and aerial imagery for detecting pedestrian infrastructure. Verma et al.~\cite{verma2024crosswalk} used YOLOv5 for crosswalk detection, achieving 71\% accuracy in Washington, D.C.\ and 89\% in Los Angeles. Antwi et al.~\cite{antwi2024automated} developed a GIS-based YOLOv2 framework that identified over 2,000 crosswalk changes in Florida. Hosseini et al.~\cite{hosseini2023mapping} created TILE2NET for generating pedestrian network datasets from aerial imagery, achieving 84.51\% mean IoU. Ning et al.~\cite{ning2022sidewalk} combined aerial and street-view imagery to link 20\% more disconnected sidewalk segments, while Moran et al.~\cite{moran2022crosswalk} assessed crosswalk distribution inequities across San Francisco neighborhoods.
 
LiDAR-based road feature detection has evolved from early geometric and elevation-based rules~\cite{clode2004automatic, wang2013automatic} to methods incorporating slope, height, and intensity features for improved robustness~\cite{li2016efficient}. Recent research integrates machine learning for segmentation, boundary extraction, and centerline fitting from point clouds~\cite{suleymanoglu20243d, wang2024automated}, and deep learning for image-based road extraction continues to be adapted to LiDAR data~\cite{chen2022road, parvathi2025automated}. Challenges remain in complex urban scenes with varying point densities and geometrically similar surfaces.
 
Early edge-based frameworks demonstrated feasibility of real-time road anomaly detection in vehicular ad hoc networks (VANETs)~\cite{bibi2021edge}. Subsequent vision-based systems advanced AI-powered defect detection and road safety assessment~\cite{paramasivam2024revolutionizing, merolla2024automated, merolla2025improving}. Multi-agent paradigms have been applied to lane detection~\cite{revilloud2016new} and UAV-based pavement monitoring with pothole recognition~\cite{silva2020architectural}, while recent multi-agent visual reasoning frameworks have demonstrated improved robustness in complex, out-of-distribution road environments~\cite{song2025multi}. These works indicate a shift from monolithic models toward agentic AI systems integrating distributed perception, collaborative reasoning, and adaptive decision-making.
 
Prior research on LiDAR-based road feature extraction and agentic AI-driven defect detection remains disconnected from accessibility-focused environmental auditing. Existing LiDAR studies emphasize geometric mapping accuracy, while agentic AI approaches primarily target vehicular safety rather than pedestrian mobility. No prior work combines high-resolution multimodal LiDAR data with agentic AI to proactively detect micro-scale barriers and excessive running slopes relevant to ADA compliance and wheelchair navigation. This highlights the lack of an accessibility-aware, agent-driven auditing system for pedestrian networks.

\color{black}

\section{Definitions, Metrics, \& Datasets}

The OmniPath system comprises multiple integrated components that collaboratively identify accessible paths for wheelchair users. It is built on a robust back-end architecture that segments road networks into smaller units and systematically evaluates each segment for potential inaccessibility. While Figure \ref{fig:systemArchitecture} presents the complete system architecture, OmniPath integrates data from multiple heterogeneous sources and performs accessibility assessment at the segment level to determine route suitability.

\subsection{Definitions of Terms}

To evaluate the pedestrian network, the system analyzes three primary geometric features that directly impact wheelchair mobility.

\subsubsection{Running Slope (Longitudinal Gradient)}
The \textit{Running Slope} is defined as the gradient of the path parallel to the direction of travel, calculated as the vertical change in elevation ($dz$) divided by the horizontal distance ($dx$) along the movement vector. This metric represents the primary exertion force required by a manual wheelchair user to ascend a path. According to ADA standards, any path with a slope exceeding 5\% (1:20) is classified as a ``ramp,'' while the absolute maximum allowable slope for any ramp is 8.33\% (1:12). Consequently, our agent flags any segment with a running slope $>8.33\%$ as a potential barrier, and those exceeding 10\% (as frequently observed in our National Mall dataset) as critical energy barriers that may be insurmountable for manual users.

\subsubsection{Cross Slope (Transverse Gradient)}
The \textit{Cross Slope} is defined as the gradient measured perpendicular to the direction of travel across the width of the pedestrian path. The ADA mandates a maximum allowable cross-slope of 2.0\% (1:48). Excessive cross-slope is arguably more detrimental than running slope, as it causes a wheelchair to drift continuously toward the curb or lower edge. This forces the user to expend asymmetric energy braking with one hand while pushing with the other to maintain a straight trajectory. Our LiDAR analysis specifically targets these ``invisible'' drifts where one side of the sidewalk has subsided, often undetected by 2D vector maps.

\subsubsection{Discontinuity (Vertical Surface Deviation)}
A \textit{Discontinuity} refers to an abrupt vertical change in surface elevation that is not a slope, such as potholes, tree root uplifts, or heaving pavement slabs. In our model, this is defined as the maximum vertical variance ($\Delta z_{max}$) detected within a single 0.5-meter segment relative to the fitted plane. ADA compliance guidelines state that changes in level between 6.4mm (1/4 inch) and 13mm (1/2 inch) must be beveled, while any vertical change greater than 13mm is strictly non-compliant. Based on these thresholds, our system flags any discontinuity exceeding 20mm (approx. 0.75 inch) as a critical ``entrapment hazard,'' representing a failure point where a wheelchair's caster wheel could become stuck, leading to a potential tip-over.

\subsubsection{Severity Classification of Audited Path Features}

To translate raw geometric metrics into actionable navigation intelligence, our agentic framework moves beyond a simple binary determination of accessibility. Instead, the system assigns a graded safety label to each 0.5 m segment based on its specific deviation from ADA standards. The severity of each sidewalk segment is quantified using a \textit{Weighted Severity Score} ($S$), which represents the cumulative impact of physical barriers on pedestrian mobility.  This calculation is performed in two stages: normalization and weighted summation.

First, to enable direct comparison across heterogeneous units, such as percentages for slopes and centimeters for vertical discontinuities, the system computes a \textit{Proportional Exceedance} ($E$). This value represents the relative extent to which a measured parameter exceeds its corresponding ADA-defined threshold ($T$). For this study, the baseline thresholds are 8.33\% for running slope, 2.0\% for cross slope, and 2.0 cm for vertical discontinuities. The exceedance is calculated as:
$$E = \max\left(0, \frac{\text{measured value} - T}{T}\right)$$

Next, the final severity score ($S$) is computed using a weighted summation of these proportional exceedance values. Each component is assigned a specific multiplier reflecting the relative biomechanical difficulty it imposes on wheelchair users:
$$S = (E_{\text{running}} \times 4.0) + (E_{\text{cross}} \times 2.5) + (E_{\text{disc}} \times 3.5)$$

The weighting factors in the severity score are intentionally asymmetric to reflect the relative biomechanical and safety risks posed by different geometric failures for wheelchair users. Running slope is assigned the highest weight (4.0) due to its direct impact on propulsion effort and forward tipping risk. Vertical discontinuities are weighted at 3.5 because of their association with caster entrapment and abrupt destabilization. Cross slope is weighted at 2.5 to capture its cumulative lateral drift effect, which increases asymmetric exertion and fatigue.

These weights are informed by ADA design guidance, empirical field observations, and prior accessibility literature. While the values may be refined through user-specific calibration, the current configuration provides a conservative and interpretable risk prioritization aligned with wheelchair mobility biomechanics.

Once the score is calculated, the segment is classified into one of four severity levels that directly inform the system's routing logic:

\textbf{Mild Violations} ($S < 0.5$) represent minor technical deviations from ADA standards, such as a running slope of 8.8\% with no other compounding issues. These are generally navigable by most users and are often imperceptible in practice, though they are documented for high-precision compliance. Consequently, the routing engine treats them as fully accessible and adds no impedance.

\textbf{Moderate Violations} ($0.5 \leq S < 1.5$) denote notable barriers that actively increase physical demand, such as a 9.5\% running slope or a moderate slope compounded by a 2.2\% cross-slope. While still navigable for power wheelchair users, manual wheelchair users will experience increased exertion and fatigue. The system marks these segments as \textit{Cautionary Zones}.

\textbf{Severe Violations} ($1.5 \leq S < 3.0$) indicate significant barriers that threaten independent mobility. Examples include running slopes exceeding 10.5\% or a steep incline paired with a vertical discontinuity. Because traversing these segments independently is highly difficult and introduces risks like lateral drifting or tipping, the routing agent assigns them high impedance, actively bypassing them unless no other viable path exists.

\textbf{Critical Violations} ($S \geq 3.0$) are extreme hazards that represent a total loss of accessibility. These insurmountable barriers such as a running slope over 11.1\% or heavily compounded failures like a 12\% slope combined with a 3.5\% cross-slope and a 2.5 cm gap, pose a high risk of forward tipping, entrapment, or equipment damage. 

\subsection{Data Sources}
To calculate the metrics defined above, the OmniPath system aggregates and analyzes data from three primary geospatial sources.

\subsubsection{DC-GIS Sidewalks Dataset}
The DC Geographic Information System (DC-GIS), managed by the Office of the Chief Technology Officer (OCTO), provides a comprehensive planimetric dataset \cite{dc_sidewalks_2023} representing all legally designated public pedestrian walkways in Washington, D.C. The dataset is derived from high-resolution orthorectified aerial imagery and is updated biennially (2015, 2017, 2019, 2021, 2023, and 2025), ensuring spatial consistency and temporal reliability.

Key attributes include \texttt{FEATURECODE}, which distinguishes between sidewalks, stairs, crosswalks, and hidden sidewalks, as well as geometric fields such as \texttt{SHAPE\_AREA} and \texttt{SHAPE\_LEN}. In this research, the DC-GIS sidewalk dataset is used as the \emph{ground truth boundary}, ensuring that all analytical agents operate strictly within legally defined public pedestrian spaces and excluding non-walkable or private surfaces.

\subsubsection{USGS 3DEP LiDAR Dataset}
The United States Geological Survey (USGS) 3D Elevation Program (3DEP) \cite{usgs_lidar_explorer_2026, usgs_downloader_2026} provides high-fidelity airborne LiDAR data required for precise surface geometry analysis. The Washington, D.C. region is typically covered by Quality Level~1 (QL1) or Quality Level~2 (QL2) data, offering point densities of $\geq 8$ pulses/m$^2$ and $\geq 2$ pulses/m$^2$, respectively, with an approximate vertical accuracy of $10$~cm RMSE$_z$.

The data are delivered in LAS~1.4 format using the NAD83 horizontal datum and NAVD88 (GEOID18) vertical datum. This dataset enables fine-grained detection of longitudinal (running slope) and lateral (cross-slope) variations essential for evaluating compliance with accessibility standards.

\subsubsection{Open Street Map Data}
OpenStreetMap (OSM)~\cite{OpenStreetMap} serves as the topological foundation for the pedestrian routing network. Relevant tags include \texttt{highway=footway} with \texttt{footway=sidewalk} or \texttt{footway=crossing}, as well as accessibility-related attributes such as \texttt{kerb=*} and \texttt{tactile\_paving}.

Although OSM provides extensive connectivity information, it often lacks reliable micro-level accessibility details. Therefore, in this study, OSM is treated as a \emph{baseline graph} that is subsequently refined and corrected through verified geometric and semantic analysis.

\section{Detailed Methodology}

\begin{figure}[h]
    \centering
    \includegraphics[width= \linewidth]{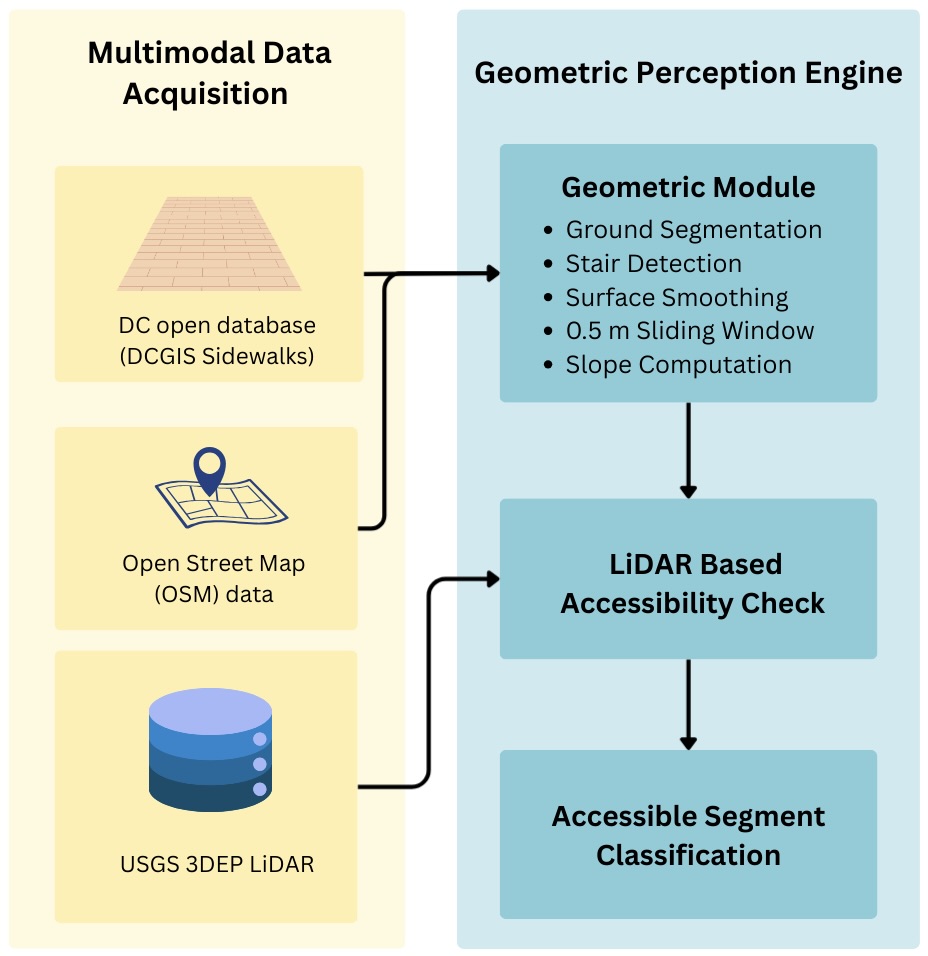}
    \caption{OmniPath System Architecture}
    \label{fig:systemArchitecture}
\end{figure}

The core objective of our framework is to transform a static, 2D map into a proactive 3D accessibility agent. To achieve this, the system must do more than simply read the map; it must physically audit the terrain. Our methodology is structured into three sequential phases. First, we merge the logical road network representing where users intend to go with high-resolution LiDAR scans that capture what they actually traverse. Second, we slice the continuous road network into discrete $0.5m$ segments, allowing the detection of localized hazards that are often invisible in standard datasets. Finally, our agent mathematically analyzes the slope and roughness of each segment, grading it against ADA standards to create a final safety map. Figure~\ref{fig:systemArchitecture} depicts the summary of our proposed methodology.

\subsection{Multi-Modal Data Acquisition and Geospatial Alignment}
To construct a high-fidelity representation of the pedestrian environment, our system fuses two distinct layers of spatial reality: the logical connectivity of the road network and the physical topography of the surface.

\textbf{1) Vector Data:} 
We established the baseline network using OpenStreetMap (OSM) data for the Washington D.C. National Mall area. This extraction resulted in 1,485 discrete pathway segments, representing sidewalks, crosswalks, and pedestrian plazas. We defined these segments as continuous paths strictly bounded by intersections or decision points. In the context of OpenStreetMap (OSM) pedestrian networks, an intersection point is a specific topological node where two or more distinct linear paths physically converge or cross. A decision point is a broader navigational concept referring to any node where a user must actively choose between multiple routing trajectories. While all intersections naturally function as decision points, decision points also include non-intersecting divergences such as path forks, crosswalk entrances, or transitions into open plazas.

\textbf{2) LiDAR Data:} 
To capture surface reality, we overlaid this network with airborne LiDAR point clouds from the 2021 District of Columbia dataset. This data provides the necessary granular detail for accessibility auditing, featuring a horizontal resolution of \SI{0.5}{m} and a vertical accuracy of $\pm \SI{5}{cm}$.


\textbf{3) Geospatial Alignment:} 
Since these datasets originate from differing coordinate systems, a \textit{Geospatial Alignment Module} standardizes all inputs into a common Projected Coordinate Reference System (CRS). This alignment is critical; it ensures that the semantic information of the road network (the ``where'') perfectly coincides with the physical surface data (the ``what'') derived from the LiDAR scans.

\subsection{The Geometric Perception Engine}

Standard navigation graphs treat a street segment as a single homogeneous edge implying that if a path is accessible at the start, it is accessible at the end. In reality, a single tree root or pothole can render an entire block impassable. To capture this reality, we deploy Agent A, our primary geometric perception unit, which audits the physical fidelity of the path at the ``step'' level.

\textbf{1) Micro-Scale Segmentation (The Wheelchair Footprint)}
Agent A begins by traversing the OSM vector network and discretizing every sidewalk edge into non-overlapping sliding windows of exactly \SI{0.5}{m}. This specific granularity was chosen to approximate the physical footprint of a standard manual wheelchair. By shrinking the analysis window from the ``block level'' to the ``wheelbase level,'' Agent A ensures that localized barriers, which would be statistically averaged out over a longer distance are isolated and exposed.

\textbf{2) LiDAR Point Cloud Association}
For each \SI{0.5}{m} window, the agent performs a spatial query to retrieve the subset of 3D LiDAR points falling within that segment's boundaries. This effectively performs a ``spatial handshake,'' associating a specific cluster of raw elevation data with a discrete logical unit of the navigation graph.

\subsection{Semantic Accessibility Auditing and Impedance Mapping}

Following the geometric quantification of the pathway, the system transitions from physical perception to semantic auditing. The Accessibility Compliance Engine ingests the raw slope and variance data derived by Agent A and evaluates it against the strict prescriptions of the Americans with Disabilities Act (ADA).  Unlike standard routing engines that view a road segment as binary (traversable or not), our auditor applies a continuous evaluation logic. It specifically isolates segments where the cross-slope exceeds the absolute maximum of \SI{2}{\%} (1:48) a critical threshold that causes dangerous lateral drift for manual wheelchair users, and flags running slopes surpassing the \SI{8.33}{\%} (1:12) limit mandated for public ramps.

To translate these diverse metrics into actionable navigation intelligence, the system maps the audited segments onto a graded severity taxonomy (see Figure~\ref{fig:OmniPathInterface}):
\begin{enumerate}
    \item \textbf{Compliant (Green):} Segments strictly adhering to federal guidelines, representing the optimal low-friction path.
    \item \textbf{Minor Violations (Yellow):} Segments with marginal deviations (e.g., Cross-Slope \SI{2}{\%}--\SI{3}{\%}). While navigable for power wheelchair users, they serve as a warning of increased physical exertion for manual users.
    \item \textbf{Critical Violations (Red):} Segments with vertical discontinuities $>\SI{2}{cm}$ or running slopes $>\SI{10}{\%}$. These are assigned a maximal impedance value, effectively pruning them from the network to ensure the agent proactively steers users away from entrapment hazards.
\end{enumerate}

\section{Performance Evaluation and Results}

In this section, we present the evaluation and efficacy of OmniPath - our proposed agentic framework. 


\subsection{Evaluation Technique}

As discussed in the previous section, OmniPath employs a sliding-window technique to detect localized barriers across the spatially aligned LiDAR and OpenStreetMap (OSM) sidewalk network prepared during the initial data fusion phase. Using a 0.5\,m sliding window with a 0.25\,m step size, this process resulted in a total of 303,361 micro-scale audit windows evaluated across the pedestrian network. A total of 1485 sidewalk segments from the National Mall area of Washington D.C., USA were analyzed using non-overlapping windows of 0.5\,m length. Of this total 1485 segments, 97 segments (6.53\%) met the requisite LiDAR point-density thresholds for reliable geometric analysis. The remaining 1,388 segments (93.47\%) were excluded due to signal attenuation caused by dense tree canopy occlusion, LiDAR flight-path gaps, or structural coverage limitations.

Despite the constrained sample size, the analysis yielded high-resolution accessibility data for LiDAR-visible corridors. Notably, 100\% of the analyzed segments contained at least one measurable ADA violation, indicating a universal prevalence of accessibility barriers within the observable network. In total, 234 critical violations (331 overall) were found among those 97 sidewalk segments (see Figure~\ref{fig:OmniPathInterface}).

To construct the elevation profile, ground-classified LiDAR points (classification code 2) falling within a 0.5 m perpendicular buffer of the path centerline are projected onto the along-path axis and binned into 10 cm intervals.  By utilizing the median elevation values within these bins, OmniPath generates a robust profile that naturally filters out noise and vegetation outliers. This continuous elevation profile then serves as the foundation for calculating key accessibility metrics. The running slope is computed via linear regression over the profile, defined mathematically as:
${slope} = \left(\frac{\Delta z}{\Delta x}\right) \times 100\%$

Simultaneously, the cross slope is determined by evaluating the elevation gradient perpendicular to the centerline. Finally, the system scans for surface discontinuities by detecting abrupt vertical elevation changes exceeding 2 cm between consecutive profile bins, effectively flagging localized tripping hazards like curbs or severe pavement uplift.

Following the micro-scale assessment, OmniPath conducts a second analysis pass using a $2.0$\,m sliding window with a $1.0$\,m stride to identify larger macro-geometric features. During this pass, the system evaluates the expanded windows for distinct structural patterns. Specifically, it looks for consecutive vertical elevation changes between $8$ and $25$\,cm coupled with tread lengths of $15$ to $45$\,cm. When these parameters are met, the features are classified as either full stairs ($\geq 5$ risers) or short stairs ($1$ to $4$ risers). Interspersed among these stair sequences, flat sections exhibiting slopes below $1\%$ are strategically flagged as pedestrian landings. Finally, the algorithm examines the overall vertical distribution of the LiDAR points; a distinct bimodal distribution in elevation is used to indicate the presence of overhead structures or concealed sidewalks, ensuring these complex 3D environments are accurately mapped for navigation.

Each segment was classified according to its maximum \textit{Weighted Severity Score} ($S$) to assess its impact on user independence and navigability. The results demonstrate a strong predominance of high-risk barriers across the analyzed network.






\subsection{Results of Evaluation}

The empirical results indicate a critical accessibility condition within the analyzed portion of the National Mall.  Out of the 97-segment subset, an overwhelming $97.9\%$ (95 segments) were classified as either \textit{Critical} or \textit{Severe}, underscoring an urgent need for targeted remediation. This high severity rating is driven by a total of 331 discrete accessibility violations detected across the subset, revealing a pronounced pattern of \textit{violation clustering} where individual paths are affected by multiple co-occurring barriers. 

On average, OmniPath identified 3.41 violations per segment. The data demonstrates highly concentrated deficiencies, with the maximum number of violations on a single segment reaching 31 instances that typically correspond to paths suffering from continuous surface degradation or complex terrain geometry. Furthermore, 75 segments, representing $77.3\%$ of the analyzed network, exhibited compound barriers. This high prevalence indicates that accessibility obstacles are rarely isolated incidents; rather, they represent systemic structural deficiencies within specific pathways.

\begin{figure*}[h]
    \centering
    \includegraphics[width= 0.9\linewidth]{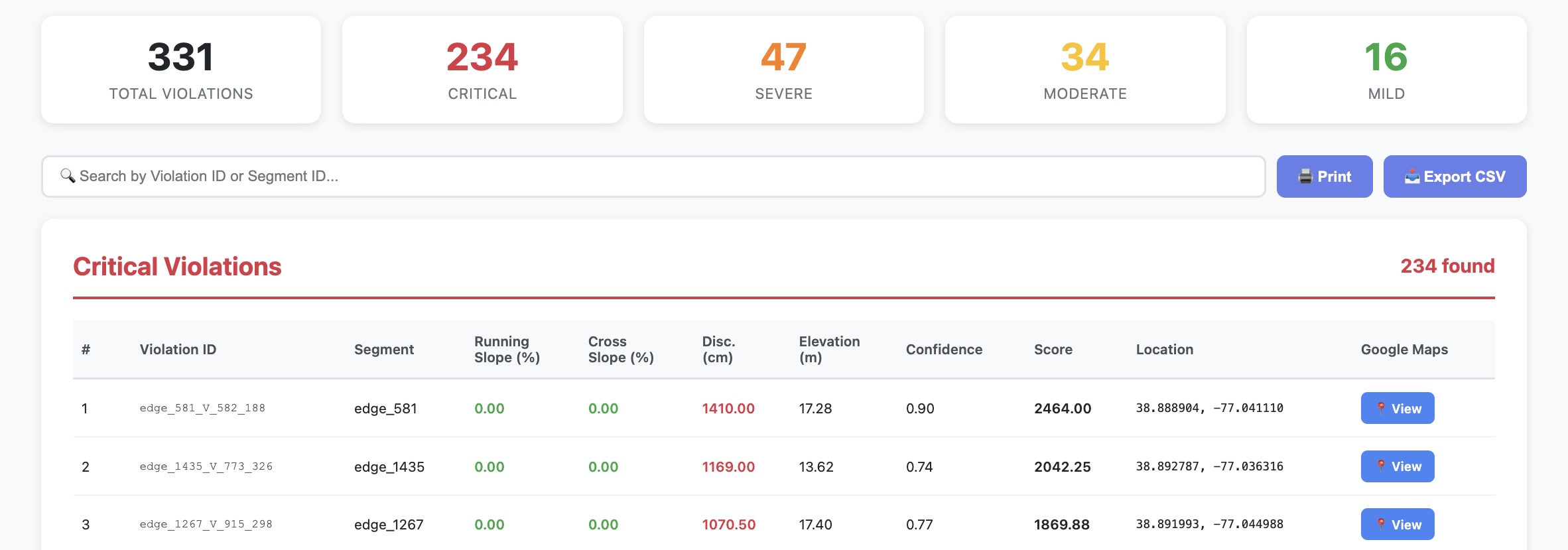}
    \caption{Interface of our OmniPath Agentic AI system}
    \label{fig:OmniPathInterface}
\end{figure*}

To facilitate the analysis and remediation of the detected barriers, the OmniPath framework includes a comprehensive web-based dashboard. As illustrated in Figure~\ref{fig:OmniPathInterface}, the interface provides a high-level summary of the 331 total violations, dynamically categorizing them across a four-tier severity taxonomy: Critical (234), Severe (47), Moderate (34), and Mild (16) violations. Beyond aggregate statistics, the system presents a detailed tabular breakdown for each individual violation. This granular view exposes the specific geometric metrics extracted by the perception engine such as running slope, cross slope, and surface discontinuities, alongside confidence scores, geographical coordinates, and direct mapping integration, thereby allowing urban planners to pinpoint and evaluate exact hazard locations.

Moving from aggregate data, Figure \ref{fig:non-accessible_sidewalks} highlights OmniPath's real-world detection capabilities. It successfully isolated specific physical hazards, including an excessive 11.76\% running slope at the Vietnam Veterans Memorial, a compound barrier (20.81\% running and 12.99\% cross slope) along Independence Avenue, and a massive 24-foot pavement discontinuity in West Potomac Park. This confirms the system's ability to accurately spot both subtle gradients and severe structural failures in the wild.

\begin{figure}[h]
    \centering
    \includegraphics[width=0.7\columnwidth]{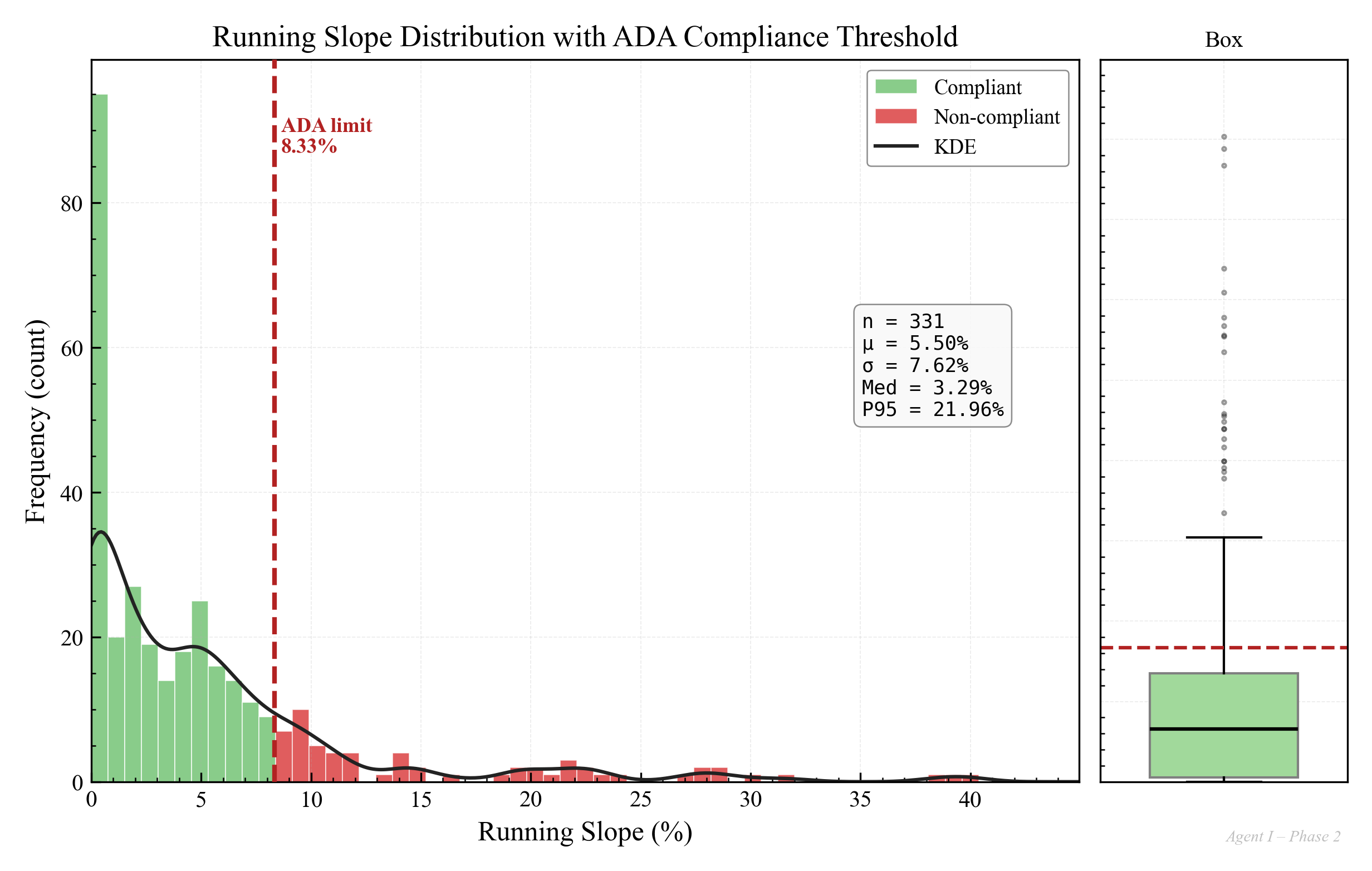}
    \caption{Running slope distribution across sidewalk segments}
    \label{fig:runningslopedist}
\end{figure}

\begin{figure}[h]
    \centering
    \includegraphics[width= 0.7\columnwidth]{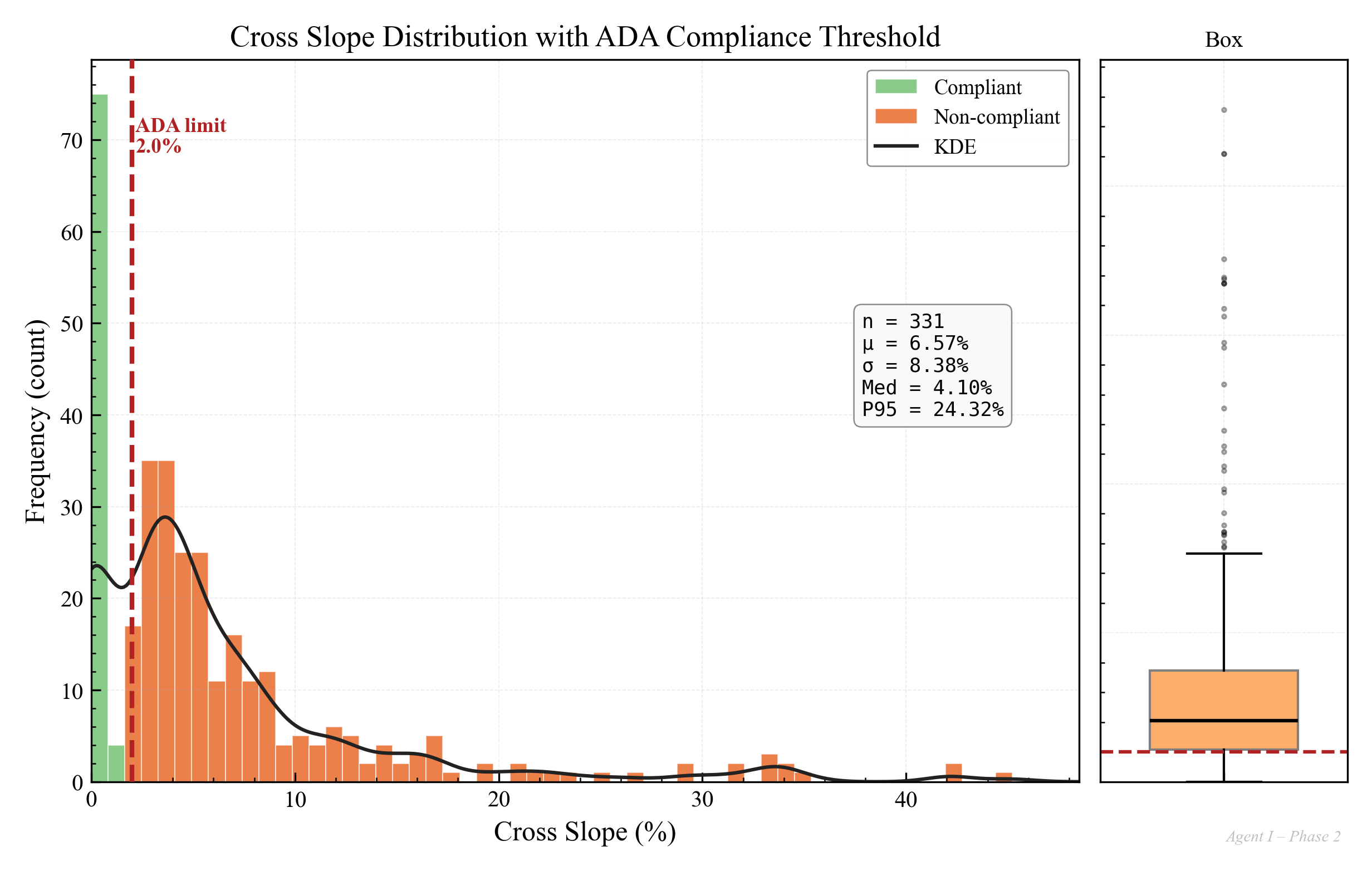}
    \caption{Cross slope distribution across sidewalk segments}
    \label{fig:crossslope}
\end{figure}

\begin{figure}[h]
    \centering
    \includegraphics[width= 0.99\columnwidth]{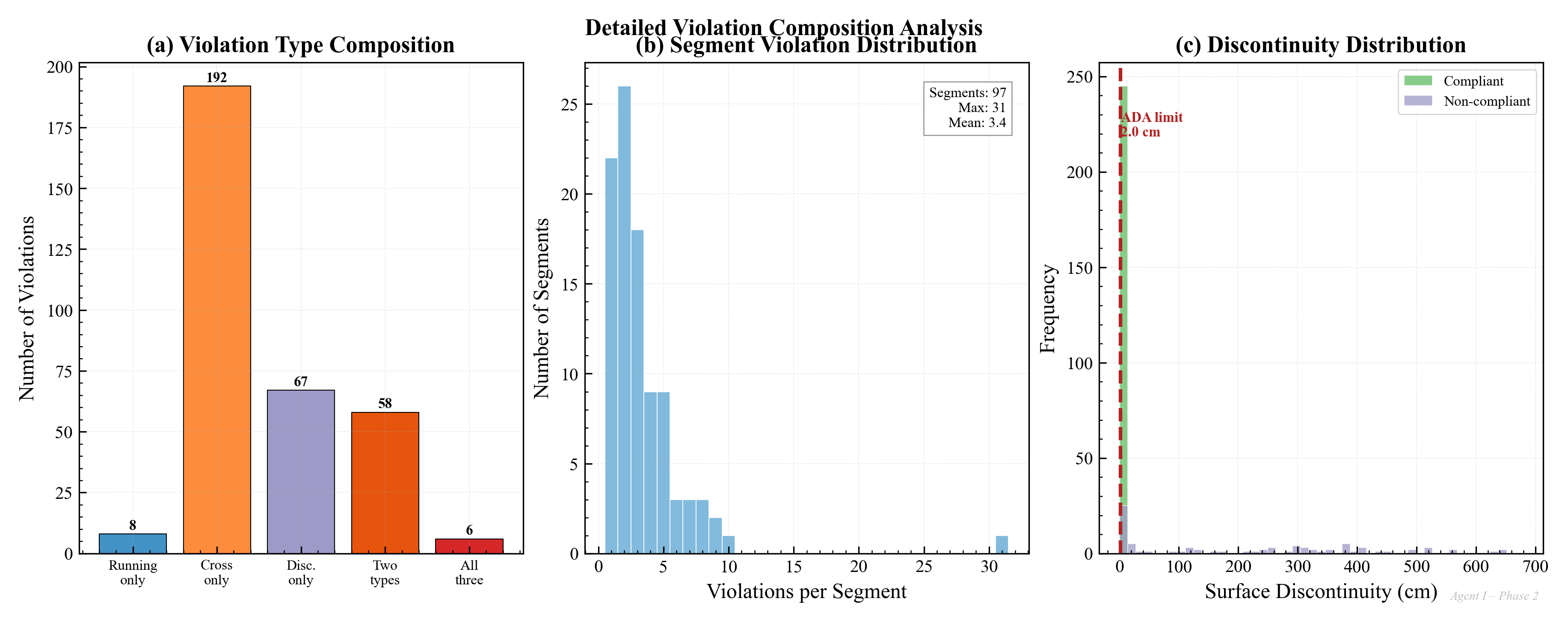}
    \caption{Detailed violation composition analysis across evaluated sidewalk segments} 
    \label{fig:compositionanalysis}
\end{figure}

Figures \ref{fig:runningslopedist} and \ref{fig:crossslope} map slope distributions against maximum ADA thresholds, while Figure \ref{fig:compositionanalysis} breaks down barrier composition. The data reveals that while many paths suffer from isolated failures, a distinct subset is heavily burdened by multiple overlapping issues. Furthermore, surface discontinuities form a long right tail in the distribution, pointing to severe, localized vertical drops rather than uniform degradation.

\begin{figure}[h]
    \centering
    \includegraphics[width= 0.99\columnwidth]{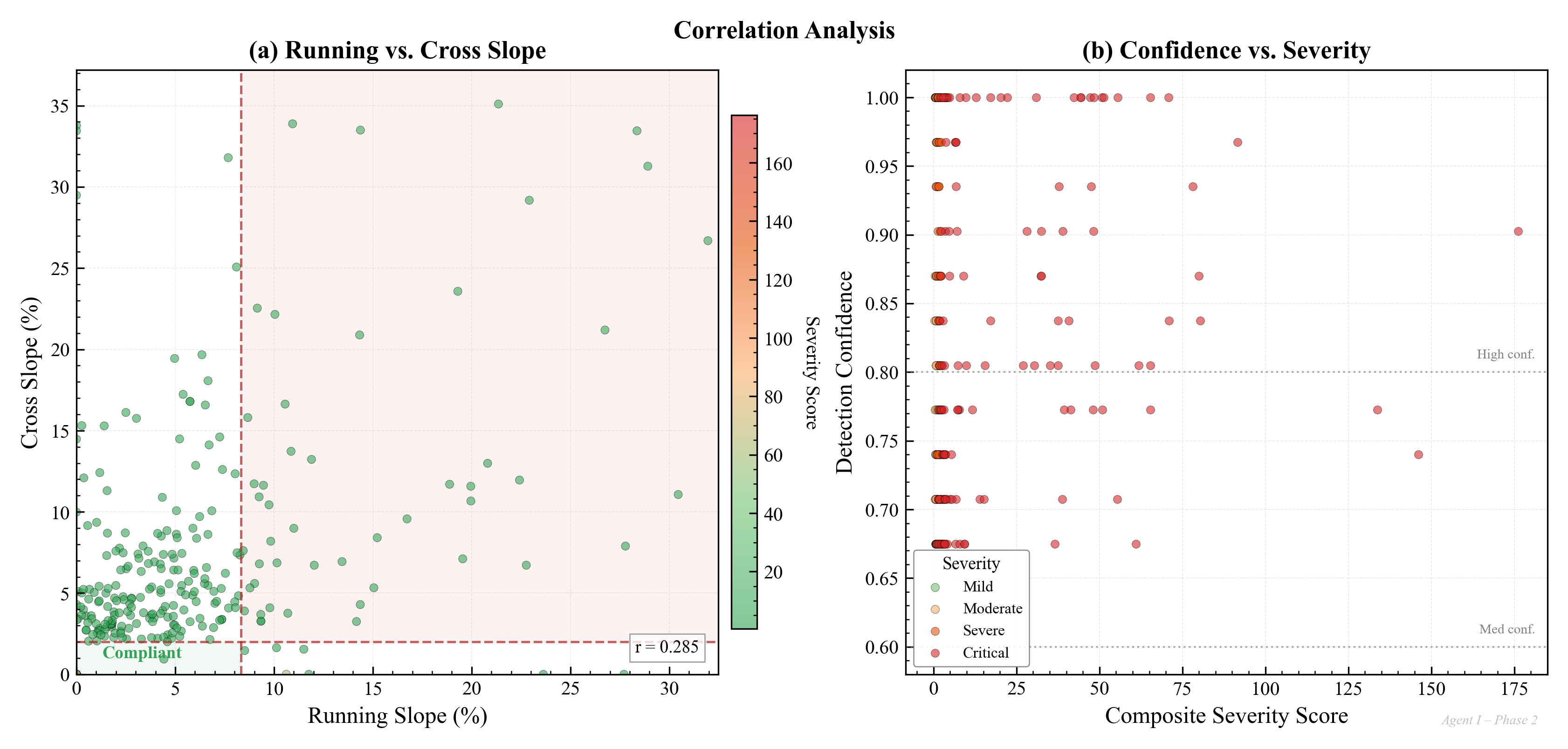}
    \caption{Correlation analysis of sidewalk accessibility metrics} 
    \label{fig:correlationanalysis}
\end{figure}

\begin{figure}[h]
    \centering
    \includegraphics[width= 0.99\columnwidth]{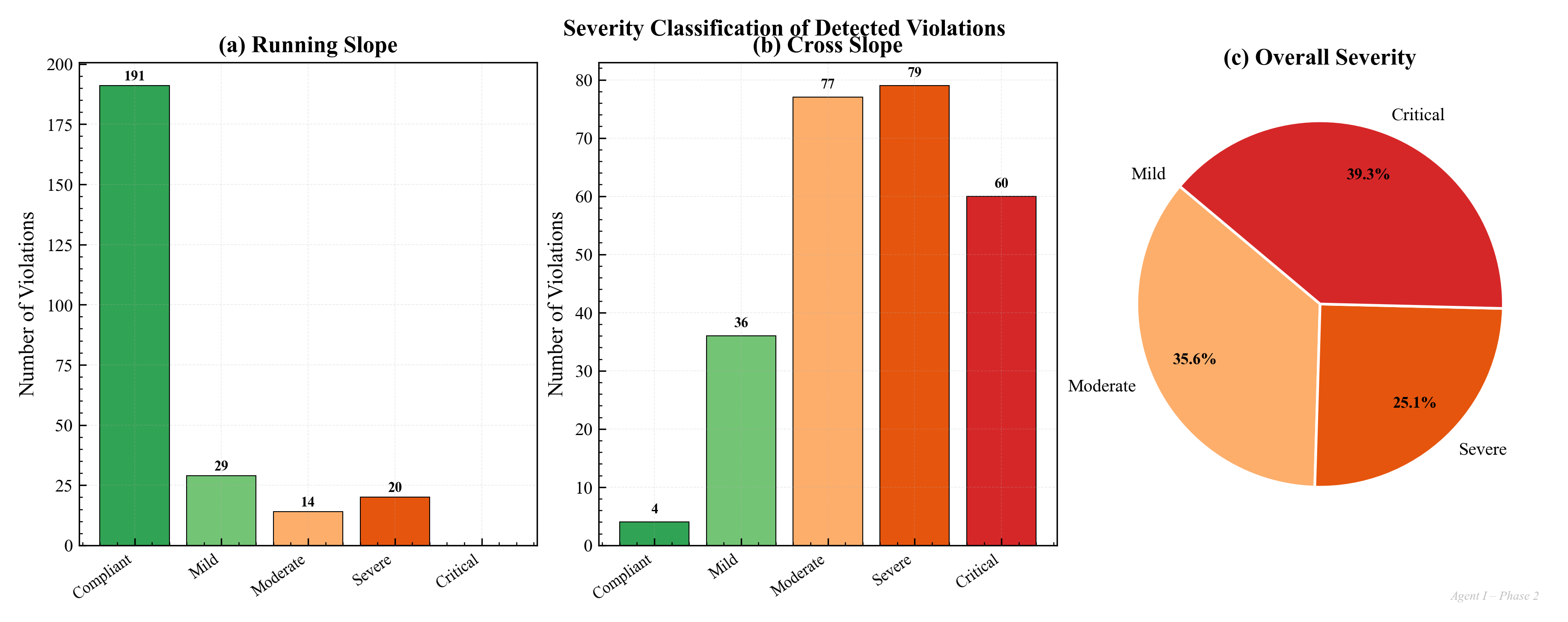}
    \caption{Severity Classification}
    \label{fig:severityclassification}
\end{figure}

Exploring the interplay of these metrics, Figure \ref{fig:correlationanalysis} shows a weak positive correlation ($r = 0.285$) between running and cross slopes, meaning dangerously steep inclines and severe sideways tilts usually occur independently. When categorized by severity (Figure \ref{fig:severityclassification}), a stark contrast emerges: nearly all running slopes are ADA-compliant, whereas cross slopes represent a pervasive network failure. The vast majority of non-compliant segments fall into the Moderate, Severe, or Critical tiers, driven almost entirely by these cross-slope deficiencies. Throughout this analysis, OmniPath consistently flagged these high-severity violations with strong model confidence.

\begin{figure}[h]
    \centering
    \includegraphics[width= \columnwidth]{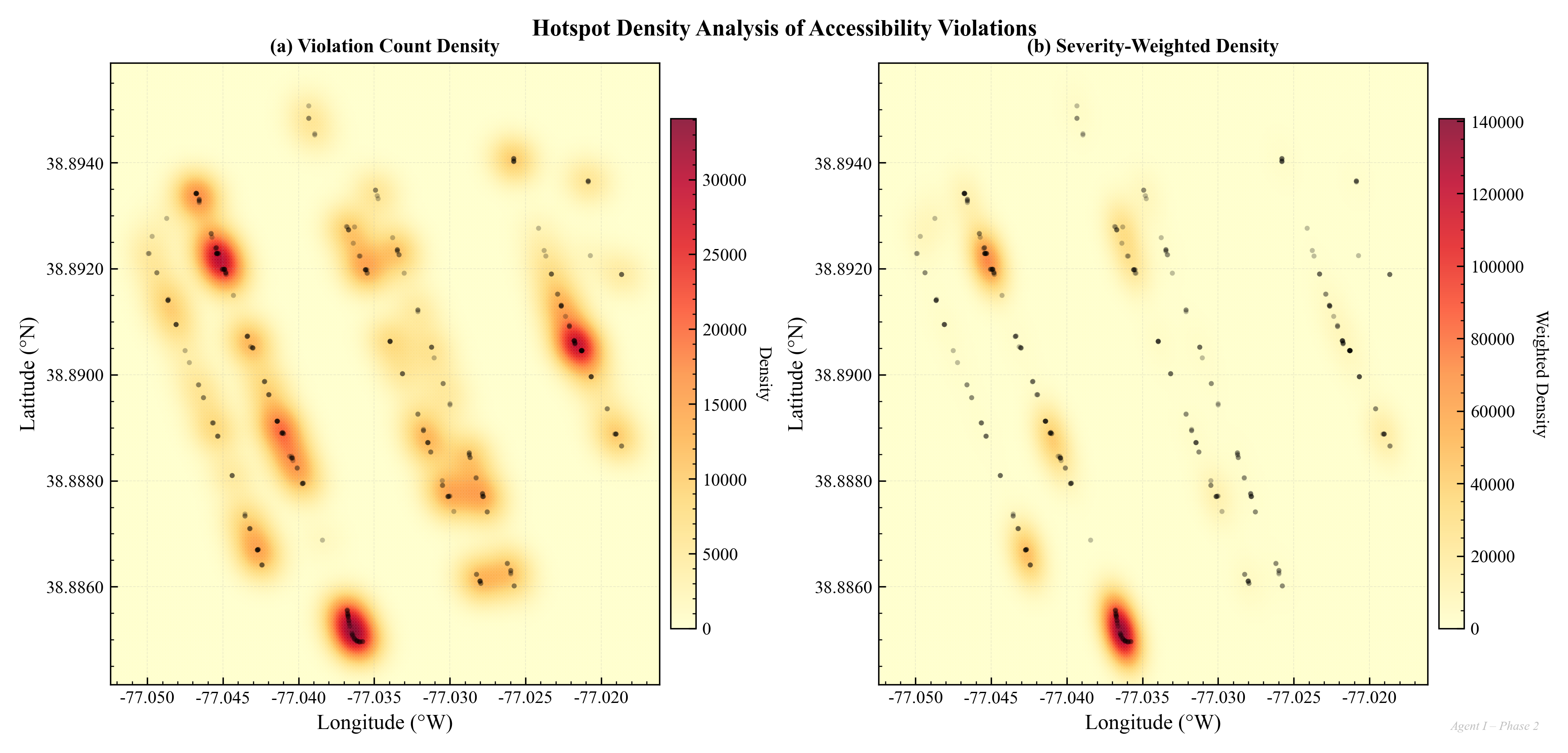}
    \caption{Hotspot Density}
    \label{fig:hotspot_density}
\end{figure}

\begin{figure}[h]
    \centering
    \includegraphics[width= \columnwidth]{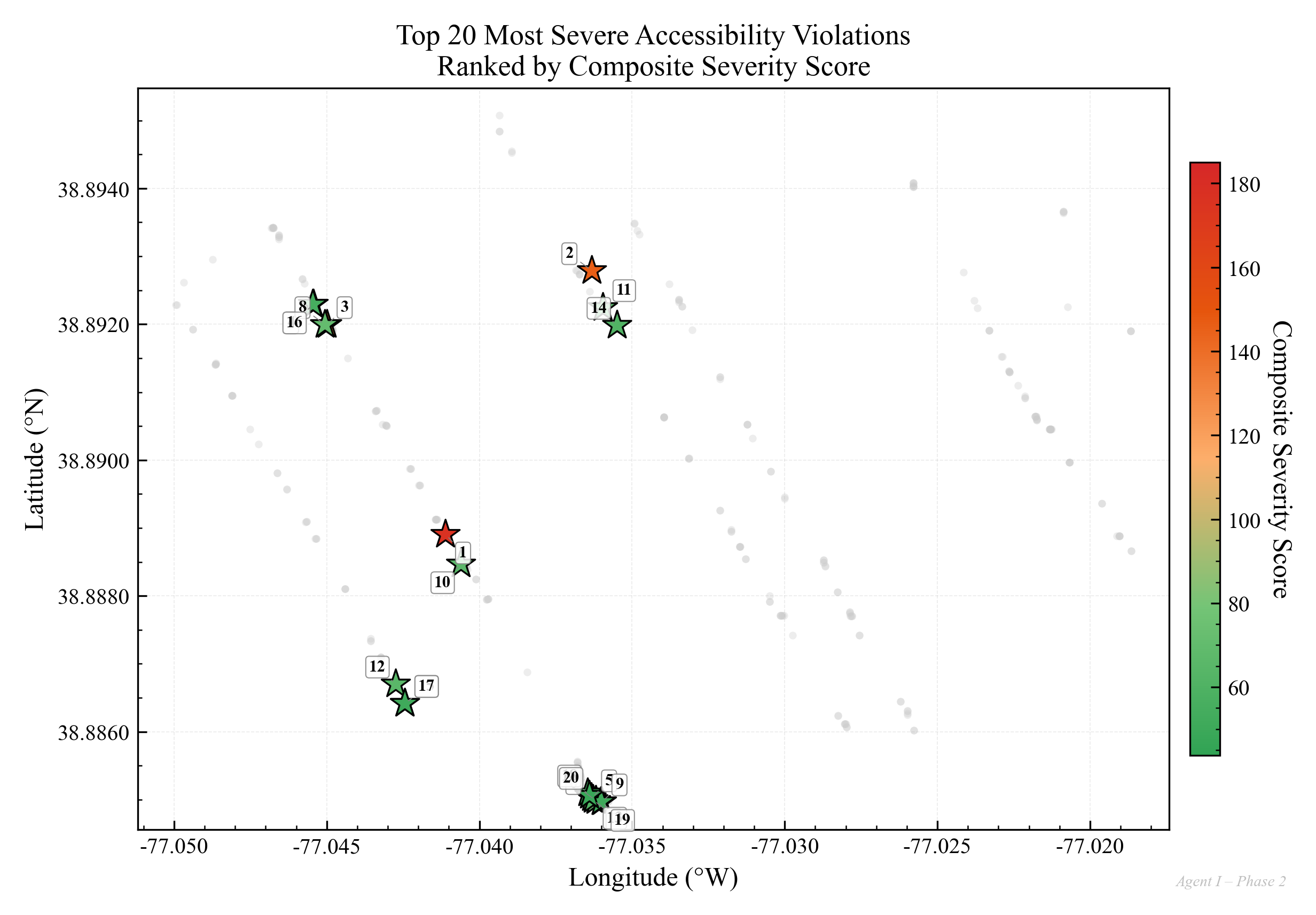}
    \caption{Spatial distribution of the top 20 most severe sidewalk accessibility violations ranked by composite severity score} 
    \label{fig:violationsmap}
\end{figure}

\begin{figure*}
    \centering
    \includegraphics[width= 0.8\linewidth]{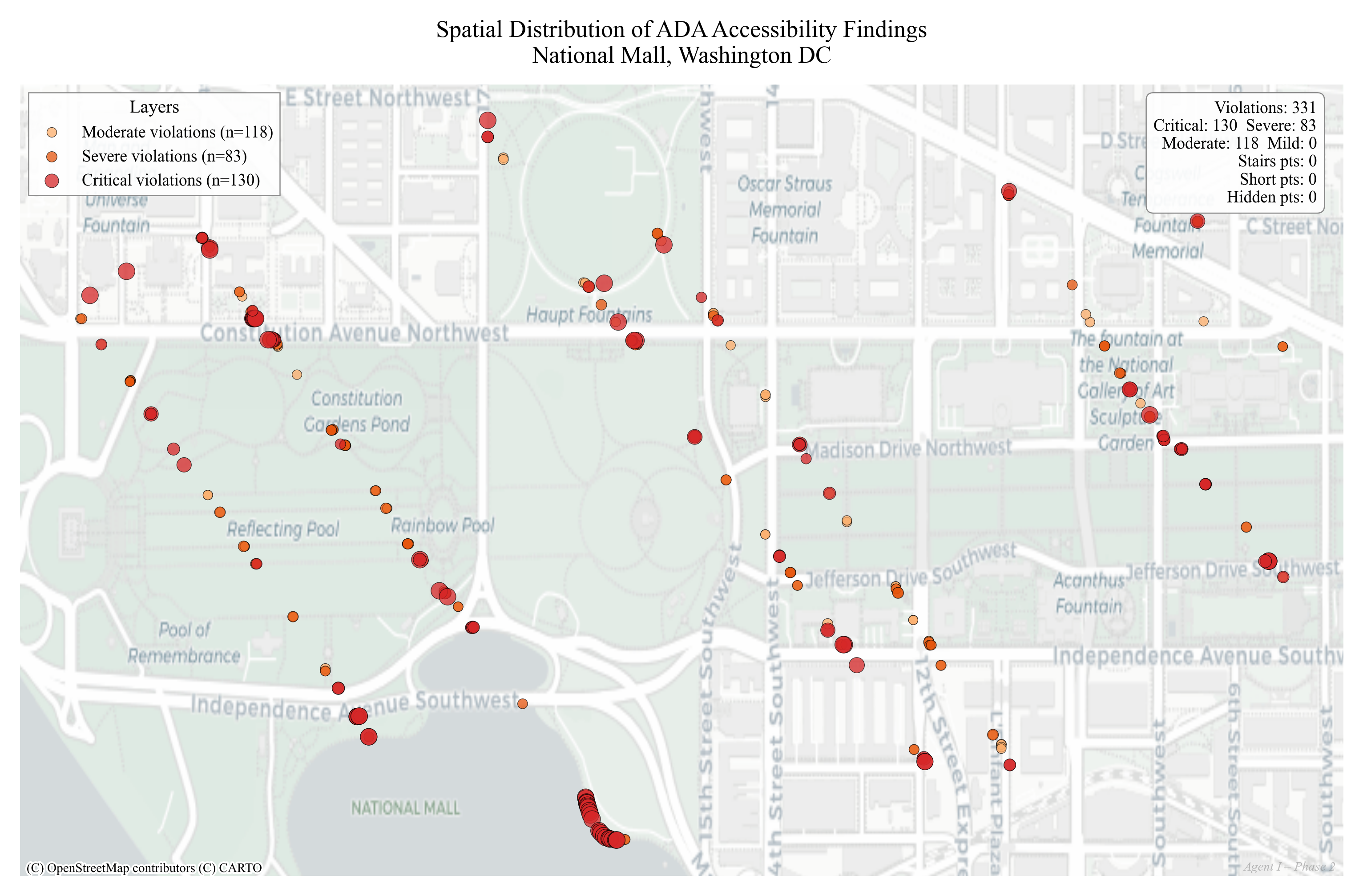}
    \caption{Spatial distribution of ADA accessibility violations across the National Mall, Washington, DC.} 
  
    \label{fig:overviewmap}
\end{figure*}

Shifting to a spatial perspective, Figures \ref{fig:hotspot_density} and \ref{fig:violationsmap} map the physical clustering of these hazards. By weighting the raw violation density by actual severity, the system isolates the truly dangerous bottlenecks, providing a prioritized roadmap of the top 20 most severe violations. Finally, pulling back to a macro-level view of the National Mall (Figure \ref{fig:overviewmap}), it becomes clear that higher-severity barriers cluster heavily along major pedestrian corridors and memorial routes. This localized clustering proves that blanket repairs would be highly inefficient, underscoring OmniPath's value in driving targeted, location-specific remediation strategies.

\subsection{Ground Truth Validation}

To rigorously evaluate the accuracy of OmniPath's agentic perception engine, we had to ensure our physical validation dataset was both statistically sound and free of sampling bias. Because the vast majority of the National Mall's sidewalk network is generally compliant, a purely random sampling approach would have overwhelmingly selected ``Mild'' segments, starving our analysis of the most dangerous hazards. To solve this class imbalance, we utilized stratified random sampling. Guided by established literature on spatial accuracy assessment, specifically Congalton's rules of thumb \cite{congalton1991review} for generating robust error matrices, we established a validation set by randomly sampling 200 pedestrian segments (roughly 15\% of the evaluated network). By ensuring an equal, proportional representation of 50 samples for each of our four severity classifications, this approach comfortably meets the baseline requirements to assume a normal statistical distribution. This guaranteed that our field team spent just as much effort verifying Critical and Severe barriers as they did baselining the Mild ones.
For the ground truth acquisition, human auditors physically visited the sampled geographic coordinates. Running and cross slopes were measured using a calibrated digital smart level , while vertical surface discontinuities were recorded using standard physical measuring tools. These field acquired measurements served as the absolute baseline to evaluate OmniPath's remote LiDAR extractions.

Beyond geometric accuracy, severity-wise evaluation (Table \ref{tab:severity_classification}) shows moderate detection capability across classes. The model performs best for Severe (F1: 0.60) and Critical (F1: 0.58) violations, followed by Moderate (F1: 0.50), while Mild cases remain challenging (F1: 0.33). Overall, the system demonstrates comparatively stronger reliability in detecting higher-severity hazards than minor deviations.

\begin{table}[htbp]
\centering
\caption{Severity-wise Binary Detection Performance}
\label{tab:severity_classification}
\begin{tabular}{@{}lccc@{}}
\toprule
\textbf{Severity Level} & \textbf{Precision} & \textbf{F1-Score} \\
\midrule
\textbf{Mild}     & 0.20  & 0.33 \\
\textbf{Moderate} & 0.33  & 0.50 \\
\textbf{Severe}   & 0.43  & 0.60 \\
\textbf{Critical} & 0.41  & 0.58 \\
\bottomrule
\end{tabular}
\end{table}

\begin{table*}[h]
\centering
\caption{Severity Distribution and Navigation Impact}
\begin{tabular}{lccc}
\hline
\textbf{Severity Level} & \textbf{Segments} & \textbf{Percentage} & \textbf{Navigation Impact} \\
\hline
Critical ($S \geq 3.0$) & 85 & 87.6\% & Insurmountable barriers; extreme risk of entrapment or injury \\
Severe ($1.5 \leq S < 3.0$) & 10 & 10.3\% & Significant barriers; requires extreme exertion or assistance \\
Moderate ($0.5 \leq S < 1.5$) & 1 & 1.0\% & Navigable but flagged as a \textit{Cautionary Zone} \\
Mild ($S < 0.5$) & 1 & 1.0\% & Minor deviation; universally traversable with minimal friction \\
\hline
\end{tabular}
\end{table*}


\section{Discussion and Conclusion}

It is important to note that the limited proportion of sidewalk segments meeting LiDAR point density requirements (6.53\%) reflects structural and environmental constraints inherent to airborne sensing in dense urban landscapes, including canopy occlusion, architectural shadowing, and LiDAR flight-path geometry. The segments with sufficient sky visibility for accurate LiDAR acquisition are predominantly located within high-traffic memorial corridors and open plazas. Consequently, accessibility conditions in heavily canopied or occluded pathways may differ from those observed within this LiDAR-visible subset.

Rather than undermining the framework’s validity, this constraint underscores OmniPath’s role as a high-precision diagnostic auditing system rather than a coarse completeness estimator. The framework prioritizes measurement fidelity over spatial coverage, ensuring that audited segments meet research-grade accuracy requirements. This design aligns with safety-critical inspection practices in civil and transportation engineering, where partial but highly reliable assessments are preferred over full-coverage estimates with uncertain error bounds. Notably, OmniPath is sensor-agnostic and can incorporate complementary data sources - such as mobile LiDAR or street-level depth sensing - to expand coverage in future deployments without altering its core agentic reasoning model.

While the current iteration of OmniPath successfully extracts high-fidelity geometric deficiencies, its reliance on purely topological LiDAR analysis presents a notable limitation regarding semantic awareness. For instance, as illustrated previously in Figure \ref{fig:non-accessible_sidewalks}(b), while the system accurately flagged the severe running and cross slopes, it could not explicitly identify the lightpost physically obstructing the pathway. To address these unclassified vertical barriers and mitigate potential false positives, future work will integrate a secondary semantic reasoning module, designated as Agent B. This multimodal decision engine will retrieve corresponding street-level imagery via the Mapillary v4 API \cite{neuhold2017mapillary} and apply a YOLOv8-based \cite{redmon2016yolo} object detection model to identify specific accessibility features, such as curb ramps, as well as physical obstructions like lightposts, parked vehicles, or trash bins. By jointly evaluating Agent A's geometric triggers with Agent B's visual semantic context, the system will be able to verify true accessibility conditions, for example, correctly reclassifying a geometrically detected ``step'' as accessible if a curb ramp is visually confirmed. Furthermore, to manage uncertainty arising from shadows, occlusions, or sparse LiDAR coverage, Agent B will compute a Bayesian confidence score combining both geometric and visual evidence; segments scoring between $0.50$ and $0.75$ will be automatically flagged for human review via the GIS dashboard, ensuring the final accessibility graph maintains research-grade reliability.

Ultimately, this study demonstrates the significant potential of the OmniPath framework as a scalable, data-driven solution for urban accessibility auditing. By successfully extracting high-fidelity geometric metrics from raw LiDAR point clouds, the system transcends the limitations of traditional, labor-intensive manual inspections. The application of this methodology to the National Mall revealed a critical landscape of clustered accessibility barriers, particularly concerning cross slopes and surface discontinuities, underscoring an urgent need for targeted infrastructure remediation. By translating these complex 3D topological deficiencies into an actionable, severity-weighted, and geolocated dashboard, OmniPath equips urban planners and policymakers with the precise spatial intelligence required to prioritize repairs efficiently. As future integrations expand its multimodal capabilities (aerial and streetview images, user reviews, etc.), OmniPath will continue to evolve into a comprehensive diagnostic tool, fundamentally advancing the pursuit of equitable and accessible pedestrian mobility in complex urban environments.

\bibliographystyle{ieeetr}
\bibliography{references}

\end{document}